\documentclass[times, 10pt,twocolumn]{article}
\usepackage{latex8}
\usepackage{times}
\usepackage{balance}
\usepackage{graphicx}

\usepackage{amsmath}
\usepackage{amssymb}
\usepackage{wrapfig}
\usepackage{multirow}

\usepackage[pagebackref=false,breaklinks=true,colorlinks,bookmarks=false,urlcolor=blue,linkcolor=blue,citecolor=blue,pdftitle={Dynamic Amelioration of Resolution Mismatches for Local Feature Based Identity Inference},pdfauthor={Yongkang Wong, Conrad Sanderson, Sandra Mau, Brian C. Lovell}]{hyperref}

\graphicspath{{./}{./figures/}}

\def\Vec#1{{\boldsymbol{#1}}}

\newcommand{\ie}{{ie.}}
\newcommand{\eg}{{eg.}}

\begin{document}

\title{Dynamic Amelioration of Resolution Mismatches\\for Local Feature Based Identity Inference}

\author
	{
	{\it Yongkang Wong, Conrad Sanderson, Sandra Mau, Brian C. Lovell}\\
  ~\\
	NICTA, PO Box 6020, St Lucia, QLD 4067, Australia~\thanks{{\bf Published~in:} \mbox{International Conf.~on Pattern Recognition (ICPR)}, pp.~1200--1203, 2010.  \href{http://dx.doi.org/10.1109/ICPR.2010.299}{http://dx.doi.org/10.1109/ICPR.2010.299} \mbox{\bf Acknowledgements:} NICTA is funded by the Australian Government as represented by the {\it Department of Broadband, Communications and the Digital Economy} as well as the Australian Research Council through the {\it ICT Centre of Excellence} program.}
  ~\\
  The University of Queensland, School of ITEE, QLD 4072, Australia
  ~\\
	}

\maketitle
\thispagestyle{empty}

\renewcommand{\baselinestretch}{0.95}\small\normalsize

\begin{abstract}
\vspace{-1ex}

\noindent
While existing face recognition systems based on~\mbox{local} features are robust to issues such as misalignment,
they can exhibit accuracy degradation when comparing images of differing resolutions.
This is common in surveillance environments
where a gallery of high resolution mugshots is compared to low resolution CCTV probe images,
or where the size of a given image is not a reliable indicator of the underlying resolution (\eg~poor optics).
To alleviate this degradation, we propose a compensation framework
which dynamically chooses the most appropriate face recognition system for a given pair of image resolutions.
This framework applies a novel resolution detection method which does not rely on the size of the input images,
but instead exploits the sensitivity of local features to resolution using a probabilistic multi-region histogram approach.
Experiments on a resolution-modified version
of the ``Labeled Faces in the Wild'' dataset show
that the proposed resolution detector frontend obtains a 99\% average accuracy
in selecting the most appropriate face recognition system,
resulting in higher overall face discrimination accuracy (across several resolutions)
compared to the individual baseline face recognition systems.

\end{abstract}
\vspace{-2ex}

\section{Introduction}
\label{sec:introduction}
\vspace{-2ex}

Face images obtained in surveillance scenarios typically have issues
such as misalignment and variations in pose and illumination.
Here we address a further issue,
namely varying image resolution~\cite{Lin_ICB_2007}, 
 encountered while undergoing real-world system trials for the UK police and other agencies.
Mismatched resolutions between probe and gallery images can cause significant performance degradation for face recognition systems,
particularly those which use high-resolution faces (\eg~mugshots or passport photos) as gallery images.
Another source of resolution mismatches is due to the fact that the size (in terms of pixels) of a given face image
may {\it not} be a reliable indicator of the underlying optical resolution.
Examples include:
{\bf (i)}~poor quality optics in low-cost cameras can act as low-pass filters;
{\bf (ii)}~poor focus and over-exposure can result in blur and loss of detail;
{\bf (iii)}~a given gallery face is provided in an already resized form and the original size is unknown
(\eg~digital scan of a photograph).

%

Face recognition approaches can be placed into two general families:
holistic and local-feature based.
In~typical holistic approaches, a single feature vector describes the entire face
and the spatial relations between face characteristics (\eg~eyes) are rigidly kept.
Examples of such systems include PCA and Fisherfaces~\cite{Belhumeur_PAMI_1997}.
In contrast, local-feature based approaches describe each face as a set of feature vectors
(with each vector describing a small part of the face),
with relaxed constraints on the spatial relations between face parts~\cite{Cardinaux_TSP_2006}.
Examples include systems based on elastic graph matching,
hidden Markov models~(HMMs)
and Gaussian mixture models~(GMMs)~\cite{Cardinaux_TSP_2006}.

Local-feature based approaches have the advantage of being considerably more robust against misalignment
(caused by automatic face detectors)
as well as variations in illumination and pose~\cite{Cardinaux_TSP_2006,Rodriguez_IVC_2006}.
As such, these systems are more suitable for dealing with faces obtained in surveillance contexts.
However, almost all of the literature on addressing resolution mismatches
(\eg~\cite{Choi_AFGR_2008,Hennings_Yeomans_CVPR_2008})
deals with holistic approaches
and naively assumes that faces are localised perfectly (\ie~no misalignment)
as well as being frontal (\ie~no pose variations).

In typical local-feature based face recognition systems,
the size of probe and gallery face images must be the same prior to feature extraction~\cite{Bowyer_TSM_2004}.
As such, the given faces are normally resized
to a common intermediate format~(IF) prior to further processing%
\footnote{{\it cf.} intermediate frequency (IF) processing in a superheterodyne radio receiver.}
 (\eg~low-resolution faces are upscaled while high-resolution faces are downscaled),
and  recognition systems are often tuned to work with that particular image size.
The use of IF processing leads to three problems in mismatched resolution comparisons, elucidated below.

\begin{figure}[!tb]
\begin{minipage}{1.0\columnwidth}
  \begin{minipage}{1.0\columnwidth}
    \begin{minipage}{0.17\columnwidth}
      \centerline{\footnotesize original}
      \vspace{-1ex}
      \centerline{\footnotesize image size}
    \end{minipage}
    \hfill
    \begin{minipage}{0.15\columnwidth}
      \centerline{\footnotesize 64~$\times$~64}
    \end{minipage}
    \hfill
    \begin{minipage}{0.15\columnwidth}
      \centerline{\footnotesize 32~$\times$~32}
    \end{minipage}
    \hfill
    \begin{minipage}{0.15\columnwidth}
      \centerline{\footnotesize 16~$\times$~16}
    \end{minipage}
    \hfill
    \begin{minipage}{0.15\columnwidth}
      \centerline{\footnotesize 8~$\times$~8}
    \end{minipage}
  \end{minipage}
  \vspace{0.5ex}
  \begin{minipage}{1.0\columnwidth}
    \begin{minipage}{0.17\columnwidth}
      \centerline{\footnotesize original}
      \vspace{-1ex}
      \centerline{\footnotesize image}
    \end{minipage}
    \hfill
    \begin{minipage}{0.15\columnwidth}
      \centerline{\includegraphics[width=1.000\columnwidth]{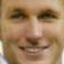}}
    \end{minipage}
    \hfill
    \begin{minipage}{0.15\columnwidth}
      \centerline{\includegraphics[width=0.500\columnwidth]{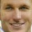}}
    \end{minipage}
    \hfill
    \begin{minipage}{0.15\columnwidth}
      \centerline{\includegraphics[width=0.250\columnwidth]{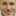}}
    \end{minipage}
    \hfill
    \begin{minipage}{0.15\columnwidth}
      \centerline{\includegraphics[width=0.125\columnwidth]{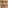}}
    \end{minipage}
  \end{minipage}
  \vspace{0.5ex}
  \begin{minipage}{1.0\columnwidth}
    \begin{minipage}{0.17\columnwidth}
      \centerline{}
    \end{minipage}
    \hfill
    \begin{minipage}{0.15\columnwidth}
      \centerline{\includegraphics[width=0.2\columnwidth]{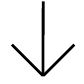}}
    \end{minipage}
    \hfill
    \begin{minipage}{0.15\columnwidth}
      \centerline{\includegraphics[width=0.2\columnwidth]{arrow.pdf}}
    \end{minipage}
    \hfill
    \begin{minipage}{0.15\columnwidth}
      \centerline{\includegraphics[width=0.2\columnwidth]{arrow.pdf}}
    \end{minipage}
    \hfill
    \begin{minipage}{0.15\columnwidth}
      \centerline{\includegraphics[width=0.2\columnwidth]{arrow.pdf}}
    \end{minipage}
  \end{minipage}
  \begin{minipage}{1.0\columnwidth}
    \begin{minipage}{0.17\columnwidth}
      \centerline{\footnotesize upscaled}
      \vspace{-1ex}
      \centerline{\footnotesize image}
    \end{minipage}
    \hfill
    \begin{minipage}{0.15\columnwidth}
      \centerline{\includegraphics[width=1.0\columnwidth]{Aaron_Peirsol_0001_64.png}}
    \end{minipage}
    \hfill
    \begin{minipage}{0.15\columnwidth}
      \centerline{\includegraphics[width=1.0\columnwidth]{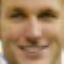}}
    \end{minipage}
    \hfill
    \begin{minipage}{0.15\columnwidth}
      \centerline{\includegraphics[width=1.0\columnwidth]{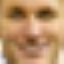}}
    \end{minipage}
    \hfill
    \begin{minipage}{0.15\columnwidth}
      \centerline{\includegraphics[width=1.0\columnwidth]{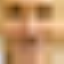}}
    \end{minipage}
  \end{minipage}
\vspace{-2ex}
\end{minipage}
\begin{minipage}{1.05\columnwidth}
  \caption
    {
    \small
    Original images of varying size upscaled to a size of {\footnotesize 64$\times$64}
    (via bilinear interpolation),
    resulting in images of fixed size but with varying underlying resolution.
    }
  \label{fig:example_inherent_res}
\end{minipage}
\vspace{-5ex}
\end{figure}

{\bf (i)} For low-resolution images, upscaling does not introduce any new information,
and can potentially introduce artifacts or noise.
Also, upscaled images are blurry (Fig.~\ref{fig:example_inherent_res}),
which causes the extracted features to be very different than those obtained from the downscaled high-resolution faces,
resulting in a significant drop in recognition accuracy~\cite{Choi_AFGR_2008}.  
Thus upscaling is generally not a good solution to the low-to-high resolution mismatch problem.
It might be tempting to employ techniques such as super-resolution or hallucination~\cite{Baker_AFGR_2000},
however super-resolution requires several images (which may not be available)
in addition to precise alignment~\cite{Lin_ICB_2007}.

{\bf (ii)} {\em Prima facie}, if upscaling is not a good solution, 
one may think that downscaling high-resolution images will solve the resolution mismatch issue.
However, downscaling reduces the amount of information available,
thereby reducing the recognition performance.
Situations can arise where the given probe face image is larger than the IF image size
(\eg~obtained through a telephoto lens).
To allow maximum accuracy wherever possible,
the recognition system should ideally be able to detect situations where a high-to-high resolution comparison is possible
(\ie~with a larger IF)
and when it should do a low-to-high resolution face comparisons (\ie~with a smaller IF).
Typically, one IF processing chain alone is not sufficient to achieve this.

{\bf (iii)} Resizing pre-supposes that the original sizes of the given images
are an indicator of the underlying resolutions.
This is often not the case in the poorly controlled image datasets encountered in practice.
Thus a resolution detector is necessary to identify whether the underlying resolution of the probe image is high or low.

In this paper we present a novel method to handle resolution mismatches
for the recently proposed Multi-Region Histograms~(MRH) local-feature approach,
which can be thought of as a hybrid between the HMM and GMM based systems~\cite{Sanderson_ICB_2009}.
Specifically, we propose:
{\bf (i)}~%
the use of two IF sizes (small and large),
with the small IF size targeted for reducing resolution mismatches caused by upscaling,
and the large IF size targeted for high discrimination performance when little to no resolution mismatches are present;
{\bf (ii)}~%
a dedicated resolution detector frontend to address situations where the actual resolution of given faces is unknown
(\ie~where the size of given faces cannot be relied upon to determine the resolution);
{\bf (iii)}~%
to employ the resolution detector,
as part of a resolution mismatch compensation framework,
to determine which of the two IF image sizes to use
when comparing two face images with unknown resolutions.

We continue the paper as follows.
In Section~\ref{sec:algorithm_MRH} we briefly describe the MRH-based face recognition approach.
The proposed resolution mismatch compensation framework is described in Section~\ref{sec:algorithm_res_det}.
Section~\ref{sec:experiments} presents experiments on the recent Labeled Faces in the Wild (LFW) dataset~\cite{LFW_techreport},
which contains problematic face variations akin to those found in surveillance scenarios.
The main findings are presented in Section~\ref{sec:conclusions}.

\vspace{-1ex}
\section{Probabilistic Multi-Region Histograms}
\label{sec:algorithm_MRH}
\vspace{-2ex}

The MRH approach is motivated by the `visual words' technique originally used in image categorisation~\cite{Nowak_ECCV_2006}.
Each face is divided into several fixed and adjacent regions,
with each region comprising a relatively large part of the face. 
For region $r$ a set of feature vectors is obtained,
\mbox{\small $F_r = \{ \Vec{f}_{r,i} \}_{i=1}^{N}$},
which are in turn attained by dividing the region into small overlapping blocks (or patches)
and extracting descriptive features from each block via 2D~DCT decomposition~\cite{Gonzales_2007}.
Each block has a size of 8$\times$8 pixels, which is the typical size used for DCT analysis.
To account for varying contrast, each block is normalised to have zero mean and unit variance.
Based on~\cite{Sanderson_ICB_2009}, coefficients from the top-left {\small 4$\times$4} sub-matrix
of the {\small 8$\times$8} DCT coefficient matrix are used,
excluding the 0-th coefficient
(which has no information due to the normalisation).

For each vector $\Vec{f}_{r,i}$ obtained from region $r$, a probabilistic histogram is computed:
\vspace{-1ex}
\begin{small}
\begin{equation}
   \Vec{h}_{r,i}
   \hspace{1pt}
   \mbox{=}
     \left[
        \hspace{-1pt}
        \frac{ w_1 ~ p_1\left(\Vec{f}_{r,i}\right)}{ \sum_{g=1}^{G} w_g p_g\left(\Vec{f}_{r,i}\right) },
        \hspace{-1pt}
       ~\cdots\hspace{-1pt},
       ~\frac{ w_G ~ p_G\left(\Vec{f}_{r,i}\right) }{ \sum_{g=1}^{G} w_g p_g\left(\Vec{f}_{r,i}\right) }
        \hspace{-1pt}
     \right]^T
     \hspace{-6pt}
\label{eqn:prob_histogram}
\vspace{-0.5ex}
\end{equation}
\end{small}%
where the $g$-th element in {\small $\Vec{h}_{r,i}$} is the posterior probability of {\small $\Vec{f}_{r,i}$}
according to the $g$-th component of a visual dictionary model.
As the visual dictionary is a mixture of Gaussians,
the mean of each Gaussian can be thought of as a particular `visual word'.

Once the histograms are computed for each feature vector from region $r$,
an~average histogram for the region is built:

\vspace{-3ex}
\begin{small}
\begin{equation}
  \Vec{h}_{r,\mathtt{avg}} = \frac{1}{N} \sum\nolimits_{i=1}^{N} \Vec{h}_{r,i}
\label{eqn:region_avg}
\end{equation}
\end{small}%

\noindent
The overlapping during feature extraction,
as well as the loss of spatial relations within each region (due to averaging),
results in robustness to translations of the face which are caused by imperfect face localisation.
The DCT decomposition acts like a low-pass filter,
with the information retained from each block being robust to small alterations
(\eg~due to minor in-plane rotations).



The normalised distance between faces {\small $X$} and {\small $Y$} is calculated using:

\vspace{-4ex}
\begin{small}
\begin{equation}
  d_\mathtt{norm} (X, Y) =
  \frac
    {
    d_\mathtt{raw}(X, Y)
    }
    {
    \frac{1}{2M}
    \sum_{i=1}^{M} \left\{ d_\mathtt{raw}(\hspace{-1pt}X, C_i\hspace{-1pt}) + d_\mathtt{raw}(Y, C_i\hspace{-1pt}) \right\}
    }
\label{eqn:norm_dist}
\end{equation}%
\end{small}%

\noindent
where {\small $C_i$} is the $i$-th cohort face and {\small $M$} is the number of cohorts,
while {\small $d_\mathtt{raw}(\cdot,\cdot)$} is a {\small $L_1$}-norm based distance measure
between histograms from {\small $R$} regions:
\begin{small}
\begin{equation}
  d_\mathtt{raw} ( X, Y ) = \frac{1}{R} \sum\nolimits_{r=1}^{R} \left\| \Vec{h}_{r,\mathtt{avg}}^{[X]} - \Vec{h}_{r,\mathtt{avg}}^{[Y]} \right\|_1
\label{eqn:raw_dist}
\end{equation}
\end{small}%
Cohort faces are assumed to be reference faces that are known not to be of persons depicted in {\small $X$} or {\small $Y$}.
The denominator in Eqn.~(\ref{eqn:norm_dist}) estimates how far away,
on average, faces {\small $X$} and {\small $Y$} are from a randomly selected face.
This typically results in Eqn.~(\ref{eqn:norm_dist}) being approximately~1
when {\small $X$} and {\small $Y$} represent faces from two different people,
and less than~1 when {\small $X$} and {\small $Y$} represent two instances of the same person.
\vspace{-1ex}
\section{Proposed Compensation Framework}
\label{sec:algorithm_res_det}
\vspace{-2ex}

In order to handle resolution mismatches when the size of given face images
cannot be relied upon as an indicator of the underlying resolution,
it is necessary to analyse the content of the given images
and determine whether the images can be downscaled to a more appropriate size.

%

In this work we use two IF image sizes, namely A and B.
We define size~A as {\small 64$\times$64}
and size~B as {\small 32$\times$32}.
It is important to note that due to the low-pass filtering effect of the DCT analysis,
MRH-based recognition tuned for size~A (where all given images are resized to size~A)
is able to handle images which have an underlying resolution ranging from {\small 32$\times$32} to {\small 64$\times$64},
while MRH-based recognition tuned for size~B is suited for {\small 32$\times$32}
and lower resolutions (\ie~{\small 16$\times$16} and {\small 8$\times$8}).

The detector uses two sets of reference faces: {\small $S_{A}$} and {\small $S_{B}$}.
In each set the faces have a canonical size of {\small 64$\times$64} pixels,
though in each set the underlying resolution is different.
Set {\small $S_{A}$} contains faces which are downscaled versions of the underlying high resolution faces.
In set {\small $S_{B}$} the underlying high resolution faces were first downscaled to {\small 16$\times$16},
followed by upscaling to the canonical size (\ie~deliberate loss of information).

The detector co-opts the framework and processing used by the MRH approach,
in order to exploit the sensitivity of local DCT features to resolution mismatches.
In essence, the detector measures whether a given face is more similar to either low-resolution or high-resolution reference faces.
The processing steps~are:

\begin{small}
\begin{enumerate}

\item
The given face {\small $Q$} is rescaled to the canonical size \mbox{({\small 64$\times$64})},
regardless of the original size of {\small $Q$}.

\item
MRH analysis with 3$\times$3 regions and 1024 visual words is performed
(using parameter settings as in~\cite{Sanderson_ICB_2009}).

\item
The average distance of {\small $Q$} to faces in sets {\small $S_{A}$} and {\small $S_{B}$} is found:
\vspace{-2ex}
\begin{small}
\begin{equation}
  d_\mathtt{avg}(Q, S_{i})
  =
  {| S_{i} |}^{-1} \sum\nolimits_{j=1}^{| S_{i} |} d_\mathtt{raw} \left( Q, S_{i,j} \right)
\vspace{-1ex}
\end{equation}%
\end{small}%
where {\small $i \in \left\{A,B\right\}$},
{\small $S_{i,j}$} is the $j$-th face of set {\small $S_{i}$}
and
{\small $| S_{i} |$} is the number of faces in set {\small $S_{i}$}.

\item
The smallest average distance,
either {\small $d_\mathtt{avg}(Q, S_A)$}
or {\small $d_\mathtt{avg}(Q, S_B)$},
indicates whether MRH tuned for either size~A or~B,
respectively, should be used for recognition.

\end{enumerate}
\end{small}


\vspace{-1ex}
\section{Experiments and Discussion}
\label{sec:experiments}
\vspace{-2ex}

We used the Labeled Faces in the Wild (LFW) dataset
which contains 13,233 face images (from 5749 unique persons)
collected from the Internet~\cite{LFW_techreport}.
The faces exhibit several compound problems such as misalignment and variations in pose, expression and illumination.
In our experiments we extracted closely cropped faces%
\footnote{Available from \href{http://itee.uq.edu.au/~conrad/lfwcrop/}{http://itee.uq.edu.au/\~{}conrad/lfwcrop/}}%
~(to exclude the background)
using a fixed bounding box placed in the same location in each LFW image.

In LFW experiment protocols
the task is to classify a pair of previously unseen faces
as either belonging to the same person or two different persons~\cite{LFW_techreport}.
Performance is indicated by the mean of the accuracies from 10 folds of the 10 sets from view~2,
in a leave-one-out cross-validation scheme
(\ie~in each fold 9 sets are used for training and 1 set for testing,
with each set having 300 same-person and 300 different-person pairs).

To study the effect of resolution mismatches,
the first image in the each pair was rescaled to {\small 64$\times$64}
while the second image was first rescaled to a size equal to or smaller than {\small 64$\times$64},
followed by upscaling to the same size as the first image
(\ie~deliberate loss of information, causing the image size to be uninformative as to the underlying resolution).
The underlying resolution of the second image varied from {\small 8$\times$8} to {\small 64$\times$64}.


In experiment 1 we evaluated the classification performance of the proposed resolution detector frontend.
Reference faces for sets {\small $S_{A}$} and {\small $S_{B}$} were taken from the training set.
Preliminary experiments indicated that using 32 faces for each reference set was sufficient.
The second image in each pair from the test set was then classified
as being suitable for MRH-based face recognition using either size~A or~B.  
Recall that an MRH-based face recognition system tuned for size~A is suited for faces
which have an underlying resolution of {\small 32$\times$32} or higher,
while a corresponding system tuned for size~B is more suited for lower resolutions.
The results, shown in Table~\ref{tab:res_detector_view2},
indicate that the frontend detector is able to assign the most suitable size almost perfectly.

\begin{table}[!tb]
  \centering
  \small
  \vspace{-1ex}
  \begin{minipage}{1.05\columnwidth}
    \caption
      {
      \small
      Classification performance of the proposed image resolution detector frontend.
      All given face images have one size \mbox{\footnotesize (64$\times$64)} but the underlying resolution varies
      (\mbox{\footnotesize 8$\times$8} to~\mbox{\footnotesize 64$\times$64}).
      Face images are classified as being suitable for MRH-based face recognition using either size~A or~B.
      MRH tuned for size~A~is suitable for images with an underlying resolution of \mbox{\footnotesize 32$\times$32} or higher,
      while MRH tuned for size~B is more suited for lower resolutions.
      }
    \label{tab:res_detector_view2}
  \end{minipage}
  \begin{tabular}{c | c c}
    \\[-4ex]
    \hline\\[-2.5ex]
    {\bf Underlying} & {\bf Size}  & {\bf Size}  \\
    {\bf Resolution} & {\bf A}     & {\bf B}     \\
    \hline \\ [-2.5ex]
    64$\times$64     &  ~99.87 \%  &  ~~~0.13 \% \\
		32$\times$32     &  ~98.06 \%  &  ~~~1.94 \% \\
		16$\times$16     & ~~~1.94 \%  &   ~98.06 \% \\
		~8$\times$8~     & ~~~0.00 \%  &   100.00 \% \\
    \hline
  \end{tabular}
  \vspace{-2ex}
\end{table}

In experiment 2 we evaluated the performance of three MRH-based systems
for classifying LFW image pairs subject to resolution mismatches.
Systems~A and~B were tuned for size~A and~B, respectively,
while the dynamic system applies the proposed compensation framework
to switch between System A and B according to the classification result of the resolution detector.

Comparing the results of the two baseline systems (A~and~B) in Table~\ref{tab:res_mismatch_view2} confirms that
System~A outperforms System~B when matching images of similar underlying resolution
(\ie~{\small 64$\times$64} and~{\small 32$\times$32}),
but significantly underperforms System~B when there is a considerable resolution mismatch ({\small 16$\times$16} or lower).
System~B is able to achieve more rounded performance
at the expense of reduced accuracy for the highest resolution ({\small 64$\times$64}).

The proposed dynamic system is able to retain the best aspect of System~A (\ie~good accuracy at the highest resolution)
with performance similar to System~B at lower resolutions.
Consequently, the dynamic system obtains the best overall performance.

We note that in three out of the four tested resolutions,
the dynamic system slightly outperforms the best underlying system.
Based on observations of the original LFW dataset, 
we conjecture that this outperformance is due to a subset of LFW images already having a low underlying resolution.

%
\vspace{-1ex}
\section{Conclusion}
\label{sec:conclusions}
\vspace{-2ex}

In this paper we have shown how 
comparing images with different underlying resolutions
can lead to a significant drop in performance for a local feature based face recognition system,
and proposed a compensation framework to improve overall performance (across several resolutions).
The proposed framework relies on a novel resolution detector
frontend which exploits the sensitivity of local features to resolution.
The performance of this resolution detection and compensation framework was demonstrated
on a resolution-modified Labeled Faces in the Wild~\cite{LFW_techreport}
dataset using the Multi-Region Histogram based recognition system.


In our experiments, two systems (A and B) were tuned to different underlying resolutions.
System~A, tuned for higher underlying resolutions, 
was shown to outperform System~B when comparing images of similar underlying resolution
({\small 64$\times$64} and {\small 32$\times$32}),
while underperforming when comparing images of very different underlying resolution ({\small 16$\times$16} and {\small 8$\times$8}).
The reverse was true for System~B, tuned for lower resolutions.
The proposed dynamic compensation framework was able to maximise performance
by applying the system best tuned for any given pair of images based on their underlying resolutions.
This potential to utilise the strengths of multiple face recognition systems clearly demonstrates the advantage of the compensation framework.

For a given pair of resolution-modified images from the LFW dataset,
the proposed resolution detector was able to classify which face recognition system was the optimal one 99\% of the time on average.
This indicates nearly perfect face recognition system selection when used in the compensation framework.


\begin{table}[!tb]
  \centering
  \small
  \vspace{-1ex}
  \begin{minipage}{1.05\columnwidth}
    \caption
      {
      \small
      Performance of three MRH-based systems
      for classifying LFW image pairs with resolution mismatches.
      All images had a fixed size of \mbox{\footnotesize 64$\times$64},
      but in each pair the second image had the underlying resolution
      varying from \mbox{\footnotesize 8$\times$8} to~\mbox{\footnotesize 64$\times$64} (see Fig.~\ref{fig:example_inherent_res}).
      Systems~A and~B were tuned for size~A and~B, respectively,
      while the dynamic system switched between system~A and~B
      \mbox{according} to the classification result of the resolution detector.
      }
    \label{tab:res_mismatch_view2}
  \end{minipage}
  \begin{tabular}{ c | c c c c }
    \\[-4ex]
    \hline \\ [-2.5ex]
    {\bf Underlying}    & {\bf System} & {\bf System} & {\bf Dynamic} \\
    {\bf Resolution}    & {\bf A}      & {\bf B}      & {\bf System}  \\
    \hline \\ [-2.5ex]
    64$\times$64        & 74.25 \%     & 70.28 \%     & 74.35 \% \\
    32$\times$32        & 70.36 \%     & 69.99 \%     & 70.47 \% \\
    16$\times$16        & 59.35 \%     & 68.08 \%     & 67.62 \% \\
    ~8$\times$8~        & 53.13 \%     & 59.40 \%     & 59.90 \% \\
    \hline \\ [-2.5ex]
    Average             & 64.27 \%     & 66.94 \%     & 68.09 \% \\
    \hline
  \end{tabular}
  \vspace{-2ex}
\end{table}

\renewcommand{\baselinestretch}{0.90}\small\normalsize
\bibliographystyle{latex8}
\bibliography{references}

\end{document}